\documentclass{article}

\usepackage[final]{neurips_2019}

\usepackage[utf8]{inputenc} % allow utf-8 input
\usepackage[T1]{fontenc}    % use 8-bit T1 fonts
\usepackage{hyperref}       % hyperlinks
\usepackage{url}            % simple URL typesetting
\usepackage{booktabs}       % professional-quality tables
\usepackage{amsfonts}       % blackboard math symbols
\usepackage{nicefrac}       % compact symbols for 1/2, etc.
\usepackage{microtype}      % microtypography
\usepackage{graphicx}
\usepackage{CJKutf8}
\usepackage{makecell}

\usepackage{CJKutf8}
\usepackage{multirow}
\usepackage{caption}

\newcommand*\rot{\rotatebox{90}}

\title{Language Transfer for Early Warning of Epidemics from 
 Social Media}

\author{%
  \textbf{Mattias Appelgren}\textsuperscript{1}, \textbf{Patrick Schrempf}\textsuperscript{1,2}, \textbf{Mat\'{u}\v{s} Falis}\textsuperscript{1}, \textbf{Satoshi Ikeda}\textsuperscript{1}, \textbf{Alison Q. O'Neil}\textsuperscript{1,3} \\
  \textsuperscript{1}Canon Medical Research Europe, \textsuperscript{2}University of St Andrews, \textsuperscript{3}University of Edinburgh\\
  \texttt{\{mattias.appelgren, patrick.schrempf, matus.falis\}@eu.medical.canon} \\
  \texttt{\{satoshi.ikeda, alison.oneil\}@eu.medical.canon} \\
}

\begin{document}

\maketitle

\begin{abstract}
Statements on social media can be analysed to identify individuals who are experiencing red flag medical symptoms, allowing early detection of the spread of disease such as influenza. Since disease does not respect cultural borders and may spread between populations speaking different languages, we would like to build multilingual models. However, the data required to train models for every language may be difficult, expensive and time-consuming to obtain, particularly for low-resource languages. Taking Japanese as our target language, we explore methods by which data in one language might be used to build models for a different language. We evaluate strategies of training on machine translated data and of zero-shot transfer through the use of multilingual models. We find that the choice of source language impacts the performance, with Chinese-Japanese being a better language pair than English-Japanese. Training on machine translated data shows promise, especially when used in conjunction with a small amount of target language data.
\end{abstract}

\section{Introduction}

The spread of influenza is a major health concern. Without appropriate preventative measures, this can escalate to an epidemic, causing high levels of mortality. A potential route to early detection is to analyse statements on social media platforms to identify individuals who have reported experiencing symptoms of the illness. These numbers can be used as a proxy to monitor the spread of the virus.

Since disease does not respect cultural borders and may spread between populations speaking different languages, we would like to build models for several languages without going through the difficult, expensive and time-consuming process of generating task-specific labelled data for each language. In this paper we explore ways of taking data and models generated in one language and transferring to other languages for which there is little or no data.

\section{Related Work}

Previously, authors have created multilingual models which should allow transfer between languages by aligning models \citep{vanderPlas:2006:FSU:1273073.1273184} or embedding spaces \citep{johnson-etal-2019-cross, alaux2018unsupervised}. An alternative is translation of a high-resource language into the target low-resource language; for instance, \citep{chaudhary2019little} combined translation with subsequent selective correction by active learning of uncertain words and phrases believed to describe entities, to create a labelled dataset for named entity recognition.

\section{MedWeb Dataset}

We use the MedWeb (``Medical Natural Language Processing for Web Document'') dataset \citep{medweb} that was provided as part of a subtask at the NTCIR-13 Conference \citep{ntcir13}. The data is summarised in Table~\ref{tab:dataset}. There are a total of 2,560 pseudo-tweets in three different languages: Japanese (ja), English (en) and Chinese (zh). These were created in Japanese and then manually translated into English and Chinese (see Figure~\ref{fig:example-tweets}). Each pseudo-tweet is labelled with a subset of the following 8 labels: influenza, diarrhoea/stomach ache, hay fever, cough/sore throat, headache, fever, runny nose, and cold. A positive label is assigned if the author (or someone they live with) has the symptom in question. As such it is more than a named entity recognition task, as can be seen in pseudo-tweet \#3 in Figure~\ref{fig:example-tweets} where the term ``flu'' is mentioned but the label is negative.

%\vspace{-2ex}
\begin{table}[!htb]
    \centering
    \caption{MedWeb dataset overview statistics.}
    \label{tab:dataset}
    \begin{tabular}{cccccccccccc}
        \textbf{Dataset} & \textbf{\rot{\# Pseudo-Tweets}} & \textbf{\rot{\makecell{Mean \# labels \\per example}}} & \textbf{\rot{Influenza}} & \textbf{\rot{Diarrhoea}} & \textbf{\rot{Hay fever}} & \textbf{\rot{Cough}} & \textbf{\rot{Headache}} & \textbf{\rot{Fever}} & \textbf{\rot{Runny nose}} & \textbf{\rot{Cold}} & \textbf{\rot{\makecell{\# Examples with \\no labels}}} \\
        \hline
        Training & 1,920 & 0.997 & 106 & 182 & 163 & 227 & 251 & 345 & 375 & 265 & 530 \\
        Test & 640 & 0.933 & 24 & 64 & 46 & 80 & 77 & 93 & 123 & 90 & 195 \\
        \hline
    \end{tabular}
\end{table}

%\vspace{-1ex}
\begin{figure}[!htb]
    \centering
    \begin{tabular}{cc|c}
        & \textbf{Pseudo-tweet} & \textbf{Labels} \\
        \hline
        (ja) & \begin{CJK}{UTF8}{min} 風邪を引くと全身がだるくなる。
        \end{CJK} & \multirow{3}{*}{Cold}\\
        (en) & The cold makes my whole body weak.& \\
        (zh) & \begin{CJK}{UTF8}{min}
        一感冒就浑身酸软无力。
        \end{CJK} &\\
        \hline
        (ja) & \begin{CJK}{UTF8}{min}
        アトピーと花粉症が重なってつらい
        \end{CJK} & Hay fever\\
        (en) & It's really bad. My eczema and allergies are acting up at the same time. & \&\\
        (zh) & \begin{CJK}{UTF8}{min}
        过敏症加花粉症，难受死了。
        \end{CJK} & Runny nose\\
        \hline
        (ja) & \begin{CJK}{UTF8}{min}
        今日インフルの手術じゃないただの注射なのにビビる
        \end{CJK} & \multirow{3}{*}{No labels}\\
        (en) & I'm so scared of today's flu shot, and it's not even surgery or anything. & \\
        (zh) & \begin{CJK}{UTF8}{min}
        今天只打针不做流感手术，但还是害怕。
        \end{CJK} & \\
    \end{tabular}
    \caption{Example pseudo-tweet triplets.}
    \label{fig:example-tweets}
\end{figure}
%\vspace{-2ex}

\section{Methods}
\label{sec:methods}

\paragraph{Bidirectional Encoder Representations from Transformers (BERT):} The BERT model \citep{devlin2018bert} base version is a 12-layer Transformer model trained on two self-supervised tasks using a large corpus of text. In the first (denoising autoencoding) task, the model must map input sentences with some words replaced with a special ``MASK'' token back to the original unmasked sentences. In the second (binary classification) task, the model is given two sentences and must predict whether or not the second sentence immediately follows the first in the corpus. The output of the final Transformer layer is passed through a logistic output layer for classification. We have used the original (English) BERT-base\footnote{PyTorch code and pre-trained models for BERT: https://github.com/huggingface/transformers}, trained on Wikipedia and books corpus \citep{bookscorpus}, and a Japanese BERT (jBERT) \citep{jbert} trained on Japanese Wikipedia. The original BERT model and jBERT use a standard sentence piece tokeniser with roughly 30,000 tokens.

\paragraph{Multilingual BERT:} Multilingual BERT (mBERT)\footnote{Multilingual BERT Models: https://github.com/google-research/bert/blob/master/multilingual.md} is a BERT model simultaneously trained on Wikipedia in 100 different languages. It makes use of a shared sentence piece tokeniser with roughly 100,000 tokens trained on the same data. This model provides state-of-the-art zero-shot transfer results on natural language inference and part-of-speech tagging tasks \citep{pires2019multilingual}. 

\paragraph{Translation:} We use two publicly available machine translation systems to provide two possible translations for each original sentence: Google's neural translation system \citep{wu2016google} via Google Cloud\footnote{Cloud Translation | Google Cloud: https://cloud.google.com/translate/}, and Amazon Translate\footnote{Amazon Translate: Neural Machine Translation: https://aws.amazon.com/translate/}. We experiment using the translations singly and together.

\paragraph{Training procedure:} Models are trained for 20 epochs, using the Adam optimiser \citep{kingma2014adam} and a cyclical learning rate \citep{smith2017cyclical} varied linearly between $5 \times 10^{-6}$ and $3 \times 10^{-5}$.

\section{Experiments}

Using the multilingual BERT model, we run three experiments as described below. The ``exact match'' metric from the original MedWeb challenge is reported, which means that all labels must be predicted correctly for a given pseudo-tweet to be considered correct; macro-averaged F1 is also reported. Each experiment is run 5 times (with different random seeds) and the mean performance is shown in Table~\ref{tab:results}. Our experiments are focused around using Japanese as the low-resource target language, with English and Chinese as the more readily available source languages.

%\vspace{-2ex}
\begin{table}[!htb]
    \centering
    \caption{Overall results, given as mean (standard deviation) of 5 runs, for different training/test data pairs. The leading results on the original challenge are shown as baselines for benchmarking purposes. EN - English, JA - Japanese, ZH  - Chinese, TJA - Translated Japanese.}
    \label{tab:results}
    \begin{tabular}{lccccc}
        \hline
        \textbf{Model} & \textbf{Source} & \textbf{Train} & \textbf{Test} & \textbf{Exact Match Accuracy} & \textbf{F1 macro} \\
        \hline
        \multicolumn{2}{l}{\textbf{Baselines}} & & & \\
        Majority class classifier & - & - & - & 0.305 \hspace{29pt} & - \\
        Random classifier & - & - & - & 0.130 (0.012) & 0.118 (0.007) \\
        \citet{naist} & - & EN & EN & 0.795 \hspace{29pt} & -  \\
        \citet{naist} & - & JA & JA & 0.825 \hspace{29pt} & -  \\
        \citet{naist} & - & ZH & ZH & 0.809 \hspace{29pt} & -  \\
        BERT & - & EN & EN & \textbf{0.847} (0.003) & \textbf{0.884} (0.004) \\
        jBERT & - & JA & JA & 0.843 (0.012) & 0.880 (0.006) \\
        mBERT & - & ZH & ZH & 0.835 (0.004) & 0.876 (0.006) \\
        \hline
        \multicolumn{2}{l}{\textbf{Zero-shot transfer}} & & & \\
        mBERT & - & EN & JA & 0.305 (0.001) & - \\
        mBERT & - & ZH & JA & \textbf{0.507} (0.007) & \textbf{0.484} (0.032) \\
        \hline
        \multicolumn{3}{l}{\textbf{Machine translation}} & & \\
        mBERT & EN & TJA & JA & 0.740 (0.011) & 0.740 (0.012) \\
        mBERT & ZH & TJA & JA & 0.774 (0.008) & 0.821 (0.010) \\
        mBERT & EN & TJA (x2) & JA & 0.754 (0.009) & 0.758 (0.034) \\
        mBERT & ZH & TJA (x2) & JA & \textbf{0.804} (0.004) & \textbf{0.849} (0.098) \\
        \hline
    \end{tabular}
\end{table}
%\vspace{-2ex}

\subsection{Baselines}

To establish a target for our transfer techniques we train and test models on a single language, i.e. English to English, Japanese to Japanese, and Chinese to Chinese. For English we use the uncased base-BERT, for Japanese we use jBERT, and for Chinese we use mBERT (since there is no Chinese-specific model available in the public domain). This last choice seems reasonable since mBERT performed similarly to the single-language models when trained and tested on the same language.

For comparison, we show the results of \cite{naist} who created the most successful model for the MedWeb challenge. Their final system was an ensemble of 120 trained models, using two architectures: a hierarchical attention network and a convolutional neural network. They exploited the fact that parallel data is available in three languages by ensuring consistency between outputs of the models in each language, giving a final exact match score of 0.880. However, for the purpose of demonstrating language transfer we report their highest single-model scores to show that our single-language models are competitive with the released results. We also show results for a majority class classifier (predicting all negative labels, see Table~\ref{tab:dataset}) and a random classifier that uses the label frequencies from the training set to randomly predict labels.

\subsection{Zero-shot transfer with multilingual pre-training}

Our first experiment investigates the zero-shot transfer ability of multilingual BERT. If mBERT has learned a shared embedding space for all languages, we would expect that if the model is fine-tuned on the English training dataset, then it should be applicable also to the Japanese dataset. To test this we have run this with both the English and Chinese training data, results are shown in Table~\ref{tab:results}. We ran additional experiments where we froze layers within BERT, but observed no improvement. 

\begin{figure}[!htb]
    \centering
    \includegraphics[width=.7\textwidth]{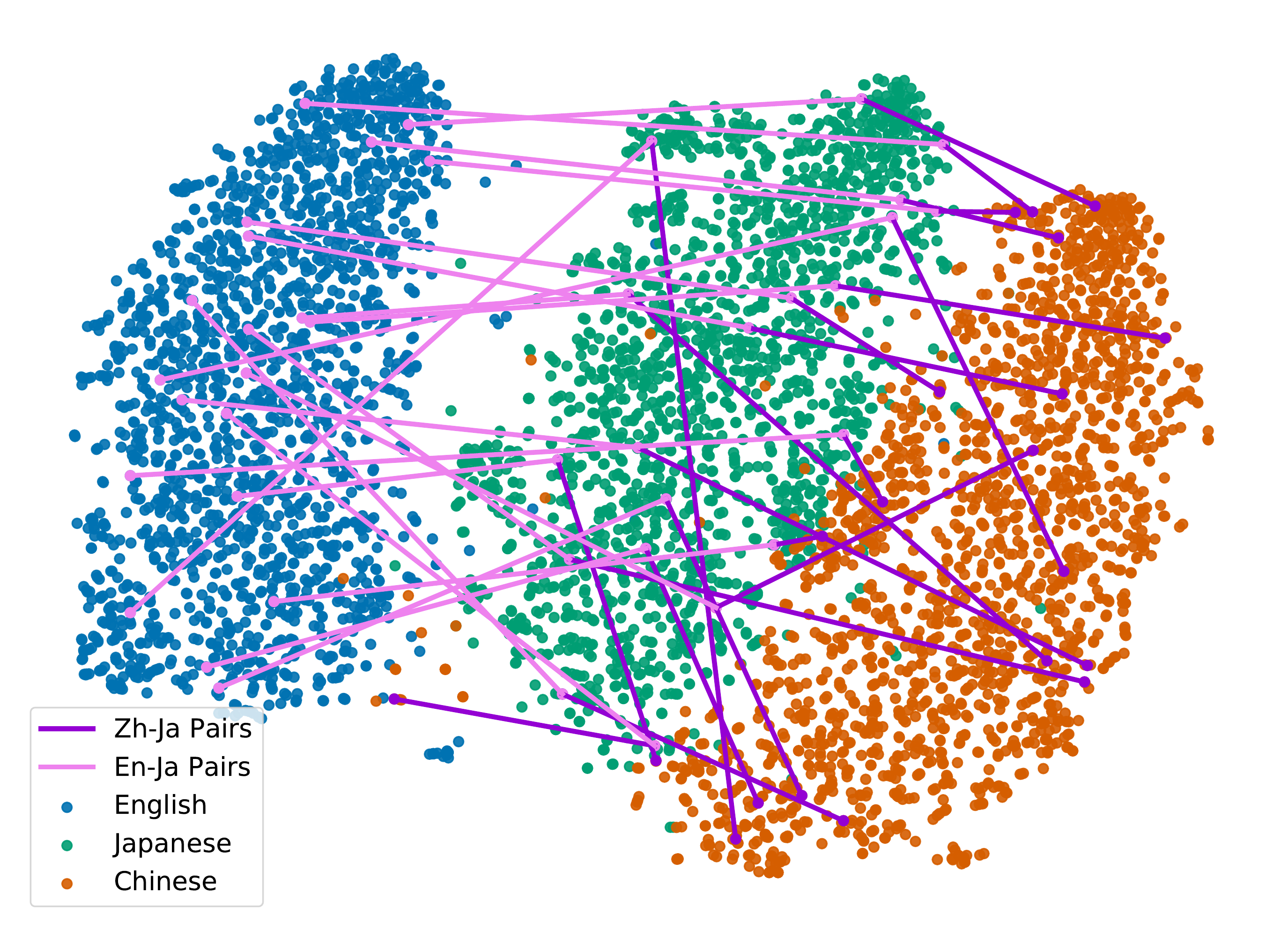}
    \captionof{figure}{Max-pooled output of mBERT final layer (before fine tuning), reduced using principal component analysis (to reduce from 768 to 50 dimensions) followed by t-SNE (to project into 2 dimensions). 20 sentence triplets are linked to give an idea of the mapping between languages.}
    \label{fig:tsne-mbert-linked}
\end{figure}

The results indicate poor transfer, especially between English and Japanese. To investigate why the model does not perform well, we visualise the output vectors of mBERT using t-SNE \citep{tsne} in Figure~\ref{fig:tsne-mbert-linked}. We can see that the language representations occupy separate parts of the representation space, with only small amounts of overlap. Further, no clear correlation can be observed between sentence pairs.

The better transfer between Chinese and Japanese likely reflects the fact that these languages share tokens; one of the Japanese alphabets (the Kanji logographic alphabet) consists of Chinese characters. There is 21\% vocabulary overlap for the training data and 19\% for the test data, whereas there is no token overlap between English and Japanese. Our finding is consistent with previous claims that token overlap impacts mBERT's transfer capability \citep{pires2019multilingual}.

\subsection{Training on machine translated data}

Our second experiment investigates the use of machine translated data for training a model. We train on the machine translated source data and test on the target test set. Results are shown in Table~\ref{tab:results}. Augmenting the data by using two sets of translations rather than one proves beneficial. In the end, the difference between training on real Japanese and training on translations from English is around 9\% while training on translations from Chinese is around 4\%.

\subsection{Mixing translated data with original data}

\begin{figure}[!htb]
    \centering
    \includegraphics[width=.8\textwidth]{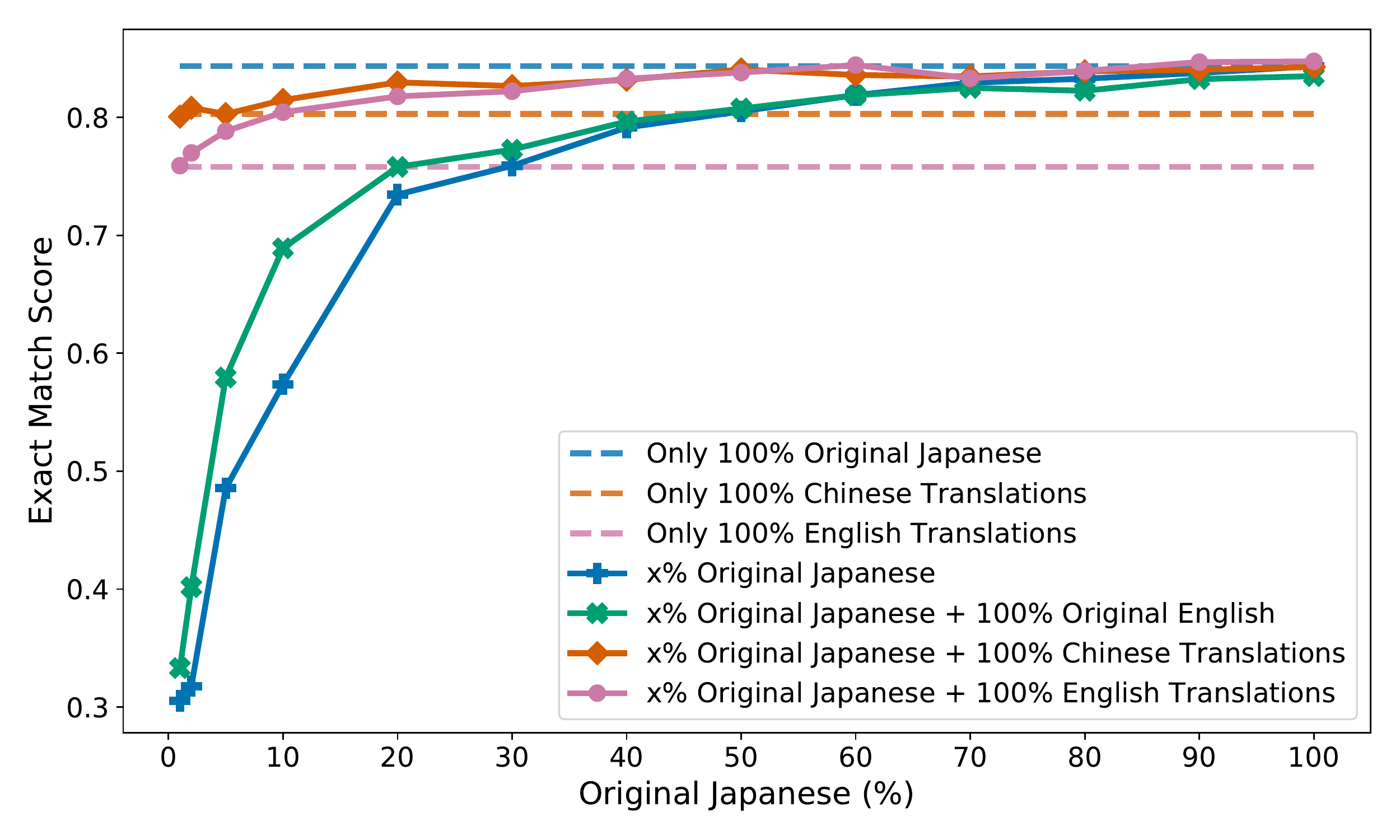}
    \captionof{figure}{Exact match accuracy when training on different proportions of the original Japanese training set, with or without either the original English data or the translated data. The pink and orange dashed lines show the accuracy of the full set of translated Japanese data (from English and Chinese respectively) and the blue dashed line shows the accuracy of the full original Japanese data.}
    \label{fig:my_label}
\end{figure}

Whilst the results for translated data are promising, we would like to bridge the gap to the performance of the original target data. Our premise is that we start with a fixed-size dataset in the source language, and we have a limited annotation budget to manually translate a proportion of this data into the target language. For this experiment we mix all the translated data with different portions of original Japanese data, varying the amount between 1\% and 100\%. The results of these experiments are shown in Figure~\ref{fig:my_label}. Using the translated data with just 10\% of the original Japanese data, we close the gap by half, with 50\% we match the single-language model, and with 100\% appear to even achieve a small improvement (for English), likely through the data augmentation provided by the translations.

\section{Discussion and Conclusions}

Zero-shot transfer using multilingual BERT performs poorly when transferring to Japanese on the MedWeb data. However, training on machine translations gives promising performance, and this performance can be increased by adding small amounts of original target data. On inspection, the drop in performance between translated and original Japanese was often a result of translations that were reasonable but not consistent with the labels. For example, when translating the first example in Figure~\ref{fig:example-tweets}, both machine translations map ``\begin{CJK}{UTF8}{min}風邪\end{CJK}'', which means cold (the illness), into ``\begin{CJK}{UTF8}{min}寒さ\end{CJK}'', which means cold (low temperature). Another example is where the Japanese pseudo-tweet ``\begin{CJK}{UTF8}{min}花粉症の時期はすごい疲れる。\end{CJK}'' was provided alongside an English pseudo-tweet ``Allergy season is so exhausting.''. Here, the Japanese word for hay fever ``\begin{CJK}{UTF8}{min}花粉症。\end{CJK}'' has been manually mapped to the less specific word ``allergies'' in English; the machine translation maps back to Japanese using the word for ``allergies'' i.e. ``\begin{CJK}{UTF8}{min}アレルギー\end{CJK}'' in the katakana alphabet (katakana is used to express words derived from foreign languages), since there is no kanji character for the concept of allergies. In future work, it would be interesting to understand how to detect such ambiguities in order to best deploy our annotation budget.

%These many-to-one relationships are indicative of the ambiguity inherent in natural language and in some cases point to subjectivity on the part of the human translators who created the dataset. 

\bibliography{main}
\bibliographystyle{plainnat}
\end{document}